# Cross-linguistic disagreement as a conflict of semantic alignment norms in multilingual AI

~Linguistic Diversity as a Problem for Philosophy, Cognitive Science, and AI~

Masaharu Mizumoto,[1] Dat Tien Nguyen,[1] Justin Sytsma,[2] Mark Alfano,[3] Yu Izumi,[4] Koji Fujita,[5] Nguyen Le Minh[1]

**Abstract**

Multilingual large language models (LLMs) face an often-overlooked challenge stemming from intrinsic semantic differences across languages. Linguistic divergence can sometimes lead to *cross-linguistic disagreements*—disagreements purely due to semantic differences about a relevant concept. This paper identifies such disagreements as conflicts between two fundamental alignment norms in multilingual LLMs—cross-linguistic consistency (CL-consistency), which seeks universal concepts across languages, and consistency with folk judgments (Folk-consistency), which respects language-specific semantic norms. Through examining responses of conversational multilingual AIs in English and Japanese with the cases used in philosophy (cases of knowledge-how attributions), this study demonstrates that even state-of-the-art LLMs provide divergent and internally inconsistent responses. Such findings reveal a novel qualitative limitation in crosslingual knowledge transfer, or *conceptual* crosslingual knowledge barriers, challenging the assumption that universal representations and cross-linguistic transfer capabilities are inherently desirable. Moreover, they reveal conflicts of alignment policies of their developers, highlighting critical normative questions for LLM researchers and developers. The implications extend beyond technical alignment challenges, raising normative, moral-political, and metaphysical questions about the ideals underlying AI development—questions that are shared with philosophers and cognitive scientists but for which no one yet has definitive answers, inviting a multidisciplinary approach to balance the practical benefits of cross-linguistic consistency and respect for linguistic diversity.

## 1. Introduction: Linguistic Diversity in Philosophy and Cognitive Science

Recent advancements in large language models (LLMs) have demonstrated remarkable capabilities in various linguistic tasks. In particular, state-of-the-art (SOTA) multilingual conversational AIs exhibit sophisticated cross-linguistic capacities, such as translation, performing the same task in multiple languages, and even generating responses in a language different from the one in which they were initially trained. The last capacity, called *crosslingual knowledge transfer*, is but one instance of the growing importance of crosslingual transfer capabilities in general.

Earlier studies on linguistic diversity using word embeddings primarily quantified

---

[1] Japan Advanced Institute of Science and Technology, [2] Victoria University of Wellington, [3] Macquarie University, [4] Nanzan University, [5] Kyoto University

semantic divergence across languages via geometric distances in representational spaces (e.g., Thompson et al., 2018; Lewis, 2023). More recent research has shifted towards improving crosslingual transfer capabilities using contextualized embeddings through LLMs (e.g., Conneau et al., 2020; Gaschi et al., 2023). These advancements aim not only to enhance practical performance but also to democratize AI technologies for low-resource languages, reflecting ethical and political considerations (Bender & Gebru, 2021).

Despite these efforts, linguistic diversity poses persistent challenges, including (1) the lack of corresponding terms for specific concepts in some languages, (2) translation inaccuracies, and (3) English-centric biases due to disparities in dataset size and quality (Bender, 2019; Joshi et al., 2020). Beyond these recognized issues, this paper introduces an additional, independent problem: what is called *cross-linguistic disagreements*, which arise from intrinsic semantic differences even when translations are accurate and alignment quality is sufficiently high. For example, basic concepts such as knowledge, intentionality, and truth may elicit different judgments across languages purely due to intrinsic semantic differences. As long as we are interested in such topics, the differences amount to disagreements between folk theories about them.

Unlike socio-cultural differences that emerge in responses to questions such as "What is the most popular sport?" and "Who is the most famous historical figure?" (Huang et al., 2024), linguistic divergence generates *conceptual* disagreements consisting of opposing but equally valid judgments across languages. Thus, while cultural alignment (Arora et al., 2022; Cao et al., 2023; Hershcovich et al., 2022) focuses on aligning AI behavior with culturally specific values and social norms, which are generally acceptable, the alignment questions that cross-linguistic disagreements raise are rather controversial. For, such disagreements challenge the very notion of truth conditions and knowledge representation. In particular, they highlight a fundamental limitation of crosslingual knowledge transfer, sometimes called crosslingual knowledge barriers (Chua, et al. 2024), as these differences can lead to opposing truth values for equivalent statements, and therefore, knowledge expressed in one language is not knowledge (being false) in another language just for the conceptual difference. Such cases then undermine the assumption that crosslingual transfer capabilities are inherently desirable and raise normative questions about how LLMs should handle linguistic diversity by revealing conflicting alignment norms.

Here, we focus on this largely overlooked issue and investigate its implications for multilingual LLMs. We begin by examining cases of cross-linguistic disagreement reported in philosophy, demonstrating their relevance not only to philosophical debates but also to cognitive science. We then investigate whether such cases also occur in state-of-the-art multilingual LLMs, highlighting the potential limitations of their crosslingual transfer capabilities, which can be called *conceptual crosslingual knowledge barriers*. We will show that such disagreements indeed occur, but there are also significant disagreements between LLMs. In the final section, we discuss normative questions (raised by these findings) regarding the conflicting ideals for training and development of multilingual LLMs. By addressing such issues, this paper aims to expand the discussion on linguistic diversity in LLMs beyond technical alignment problems, calling attention to the philosophical and normative dimensions of this increasingly important field of research, which are currently mostly about how novel techniques improve their behaviors while we question the assumption behind the very notion of "improvement" there.

**Methodological Notes:**
Unlike previous large-scale studies focusing on quantitative aspects of cross-linguistic semantic diversity or directly investigating the internal representations of LLMs layer by layer (with methods such as cosine similarity analysis or Procrustes analysis), this study adopts a qualitative approach with philosophical considerations and examines the responses of LLMs to the cases used in philosophy. This is because, on the one hand, even cross-linguistically consistent responses do not assure cross-linguistically universal representations in intermediate layers (Deb, et al. 2021), and conversely, we cannot naively assume that cross-linguistically consistent responses follow from universal representations (because of, say, the task head) either (Huang, et al. 2024). In this sense, the representations in intermediate layers (whether similar or different) do not necessarily predict the outputs.

Humans usually do not have access to their own brain states or internal representations, and philosophers have rather used the method of cases. The situation is no different for LLMs in this respect. Although there is literature on the introspective capacities of LLMs (Li, et al., 2024; Renze & Guven, 2024; Wang, et al., 2024; Yan, et al., 2024), they are mainly concerned with the capacity of LLMs to utilize their own potential or actual outputs, not the one to have direct access to their own internal representations.

Besides, even if an LLM has developed a single common representation of a concept across languages, it is very likely that the "universal" inner language rather reflects the overwhelming amount of *English* data.[1] Also, according to Zhao, et al. (2024), in the intermediate layers (task-solving layers) of contemporary LLMs, English-centric processing takes place.[2] If so, such internal common representations are not really language-neutral after all.

Indeed, if we are concerned with concepts and meanings, internal representations are not so important. This is because meaning is arguably *normative* (Boghossian, 2005; Gibbard, 1994; Kripke, 1982). For example, many philosophers agree that mere dispositions cannot constitute meaning (cf. Kripke, 1982). One can judge one's own dispositions to be wrong and try to correct them. Thus, if an LLM has a cross-linguistically common representation of some concept (in a penultimate layer) but nevertheless gives different responses by language (due to the final layer and/or the task head), we may take it as following specific (local) linguistic norms, despite its dispositions to do otherwise. This is also why we use here commercial open-access multilingual (COM) conversational AIs, which reflect the policies and ideals of their developers.

**1-1: Linguistic Divergence as a Problem in Philosophy**
Suppose that it is correct to say in one language that someone in some situation knows something, but it is also correct to say in another language that the same person in the same situation *does not* know it. If this happens, the speakers of the two languages disagree with each other in knowledge attributions purely due to linguistic differences.

---

[1] As of May 2024, English constitutes 50.5% of all websites on the internet. See: https://w3techs.com/technologies/overview/content_language..

[2] There, non-English inputs are converted into English, and reasoning is performed based on English before back-translated into the original language.

This kind of case has been reported by a group of Japanese philosophers. In particular, this includes data reported on linguistic diversity between English and Japanese with regard to philosophically important concepts, such as propositional knowledge (Mizumoto 2018, 2021), knowledge-how (Mizumoto et al., 2020; Tsugita et al., 2022), and even possibly truth (Mizumoto, 2022, 2024b).[3] We will focus on one specific example here—the concept of knowledge-how, following the recent report by Mizumto (2024a), whose data shows particularly large differences in knowledge-how attributions between English speakers and Japanese speakers.

Knowledge-how has been a traditional topic in philosophy since the last century (Ryle 1946, 1949). Intellectualists claim that knowledge-how is a species of knowledge-that (propositional knowledge), while anti-intellectualists claim that knowledge-how cannot be reduced to knowledge-that. More recently, some intellectualists gave a linguistic argument for intellectualism, based on the syntactic uniformity of constructions for knowledge-wh (knowledge with question words, such as "knows what," "knows who," etc.), in which the knowing how construction is included (Stanley and Williamson 2001, Stanley 2011a, 2011b). However, since this argument relies on *English* syntactic structures, knowing how constructions in various languages have been examined and discussed (e.g., Rumfitt 2003, Stanley 2011b, Wiggins 2012, Ditter 2016, Douskos 2013).

The assumption underlying the controversy is that there is a state of knowledge-how, which is one and the same state captured in various linguistic constructions across a range of languages. However, the nature of the debate will be very different if there is a radical semantic difference, even in extension, between different knowing how constructions in different languages.

In Mizumoto (2024a), two cases were presented to ordinary English speakers and Japanese speakers and they were asked whether the protagonist have the relevant knowledge-how. Consider the "Ski Case", in which a protagonist who has never skied before learns how to ski by reading textbooks and watching videos. Empirical results indicate that that the majority of Japanese speakers judge the protagonist to know how to ski (68%, 95% Confidence Interval: 58-76%) while most English speakers deny this (19%, 95%CI: 13-28%).

---

[3] Such data are particularly philosophically significant since philosophers have debated whether there are significant demographic differences in folk intuitions about philosophical topics (e.g., Knobe 2019, 2023, Stich and Machery 2023). However, what both parties did not consider was the possible difference in philosophical intuitions arising from linguistic diversity of philosophical concepts. The reason for this unfairly skewed attention is that participants in the debate are implicitly concerned with whether such intuitions are innate (cf. Knobe 2023, p. 408).

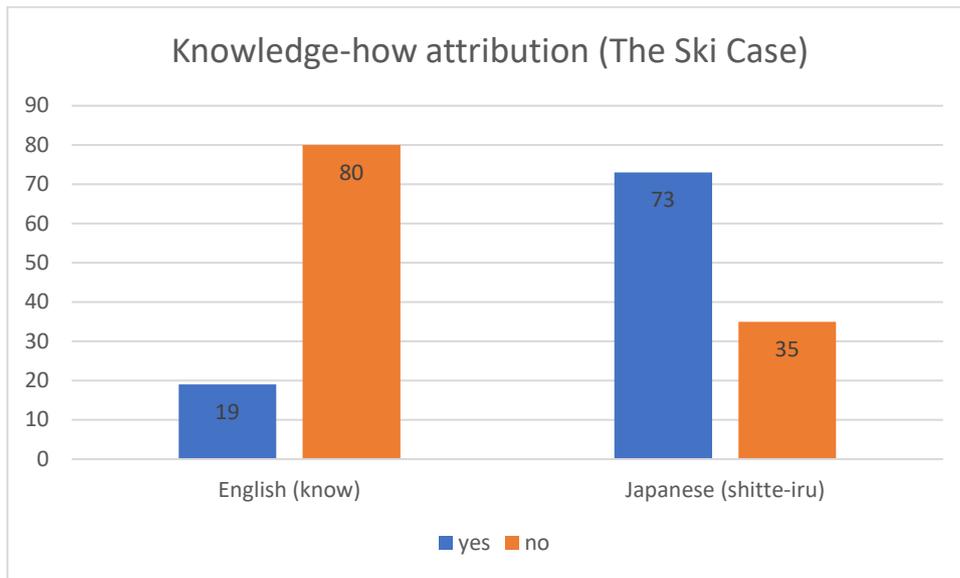

<Figure 1 reconstructed from the data of Mizumoto (2024a)>

Or consider the "Karaoke Case": the protagonist acquires an ability to sing well through an artificial means without any training, and now believes that one can sing well. This is still judged to be the case of knowledge-how by the vast majority of English speakers (79%, 95% Confidence Interval: 69-86%) but not by the vast majority of Japanese speakers (17%, 95%CI: 10-25%).

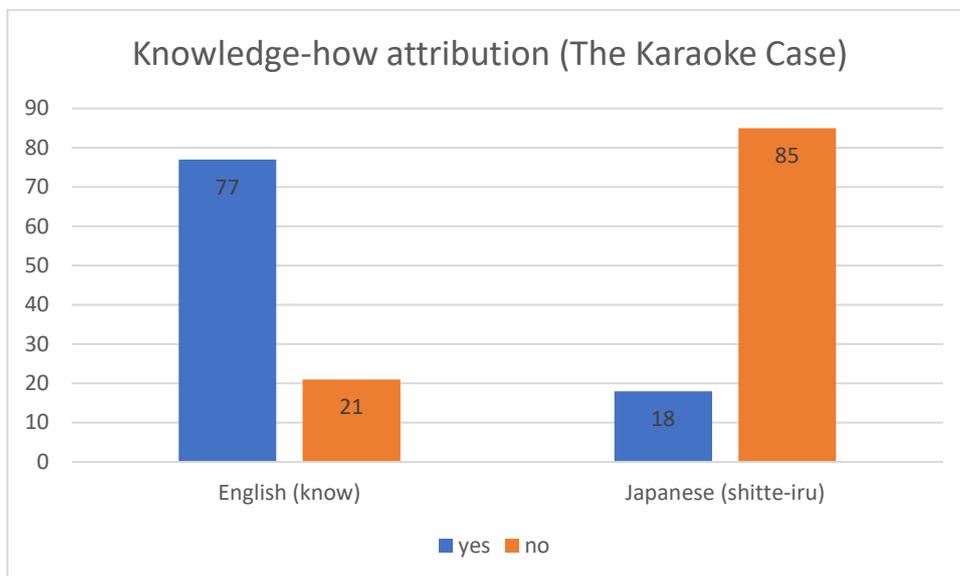

<Figure 2 reconstructed from the data of Mizumoto (2024a)>

These are stunning differences, and they are even surprising for Japanese speakers, including Japanese philosophers.

Even though not so dramatic as the results just noted, basically the same findings have repeatedly been observed in earlier studies (Mizumoto, et al. 2020, Tsugita et al. 2021). Further, Izumi et al. (2019) linguistically analyzed Japanese knowing how constructions, finding that the relevant modality associated with Japanese knowing how constructions is deontic, rather than dispositional, which also predicts the above results.

As discussed in Mizumoto (2024a), in the absence of known (radical) cultural-psychological differences between English speakers (Americans, in this case) and Japanese—which would require two, rather than one, large effects in the opposite directions—the differences may safely be taken to be linguistic in nature. (Here we can assume that in the absence of such radical cultural-psychological differences, no plausible pragmatic explanations are available either.) The present differences should therefore be explained by the difference in the *concept* of knowledge-how captured by respective languages. However, precisely because the differences are conceptual and large, differences in the *extension* of the notion of knowledge-how also follow (unless one is willing to endorse an implausibly massive error theory concerning one of the linguistic communities), and with it corresponding differences with regard to *truth conditions*.

This is not a paradoxical, let alone contradictory, state of affairs. It is rather unlikely that the concepts captured by different languages are always identical in all respects. This possibility of multiple concepts about the same topic or subject matter is called *pluralism* (or *linguistic pluralism* in this context; Mizumoto, 2022). Anglophone philosophers tend to deny this idea, treating it as a kind of relativism, but it should be seen as an open empirical possibility that needs to be tested empirically and that cannot reasonably be denied *a priori*.

At this point, one might take knowledge-how as a technical notion, which is unique and normative, rather than a folk concept that is contingent and can show linguistic divergence. It is this technical notion that is under philosophical and scientific investigation, one might say. However, since this technical notion is based on the English concept, *another* technical notion can be defined based on the Japanese concept, and if so, even the corresponding scientific investigation of it is equally possible. Such linguistic diversity, therefore, poses a problem for cognitive science if cognitive scientists are to investigate knowledge-how empirically.

### 1-2: Cognitive Science and Linguistic Diversity

Most philosophers would say that the question of knowledge-how is independent of any particular language, being about the *state* of knowledge-how. It is therefore a topic to be investigated empirically together with cognitive scientists. Thus, even Jason Stanley, who launched the linguistic argument for intellectualism, still agrees with such a conception of the topic, although he argues that "it could hardly be that science could discover that knowing how to swim was a distinct state than is expressed by 'knowing how to swim'" (Stanley, 2011a, p. 144). For, according to him,

> Cognitive scientists are unprepared simply to jettison the folk notions of knowing how and knowing that. In fact, they seem to take the ordinary notions as guiding their inquiry. (*ibid*. p. 148)

Some cognitive scientists may not agree with this claim, but part of the reason Stanley provides to support his claim is that the following sentence is "widely considered to be an a priori truth" (*ibid*. 144):

"Ana knows how to swim" is true if and only if Ana knows how to swim.

If cognitive scientists empirically investigate knowledge-how, they will be able to determine whether Anna knows how to swim (as long as the relevant data is available), which will then determine whether "Anna knows how to swim" is true or not. However, if the above bi-conditional is translated into Japanese (which should also be true *a priori* if the bi-conditional in English is) and Japanese cognitive scientists empirically investigate when or in what condition Anna knows how to swim, then the study would end up with a very different truth condition (on the right-hand side of the bi-conditional) than that of the Anglophone cognitive scientists.

More specifically, cognitive scientists may investigate whether the intellectual (propositional) cognitive component and brute-causal motor processes are independent (Stanley & Krakauer, 2013), or whether the latter (motor control systems, motor acuity, motor representations, etc.) are themselves intelligent (i.e., play a crucial role in realizing intelligence) (Fridland, 2017; Levy, 2017). Having a very different concept of knowledge-how would affect the answer to such a question, even if there is no disagreement over the empirical facts. Indeed, hypothetical monolingual Japanese cognitive scientists would not investigate motor processes in their investigations into knowledge-how in the first place. Even if the object of the study is conceived as *skills*, which are called "waza" in Japanese (see for *waza*, Ikuta 1987), it is not usually considered a species of intelligence, let alone knowledge, but a non-mental physical ability (note that mere involvement of brain processes does not necessarily make the process as a whole *mental*). From this perspective, even Anglophone anti-intellectualists are too intellectualistic for the Japanese.

Beyond the question of whether knowledge-how is a species of knowledge-that, cognitive scientists have also tried to define the notion of the automatic process (Bargh, 1994; Moors, 2016; Rawson, 2010). To explain it, the concept of "caching"—analogous to cache memory in computers—has been proposed (Cushman & Morris, 2015; Haith & Krakauer, 2018; Huberdeau et al., 2019). There, caching memory of SR (stimulus and response) associations is supposed to reduce computational costs. However, some cognitive scientists further claim that a cached movement policy is a form of procedural knowledge (Haith & Krakauer, 2018) or a mental algorithm, which is not free from controversy (Fresco et al., 2023). Even if such a function is physically realized, whether it is a form of knowledge or even a *mental* state (and whether the consequent motor control system is *intelligent*) can be a question beyond the details of the mechanism (and the question itself might have stemmed from the assumption that it is part of knowledge-how). However sophisticated, if a motor process is automatic, whether to consider it as a *mental* process is an option and can depend on the researchers' metalanguage.

In fact, this kind of worry has been expressed in linguistics by lexical semanticists. Thus, according to advocates of Natural Semantic Metalanguage (NSM),

> Anglophone scholars in the human sciences often unwittingly frame their research hypotheses in English-specific terms. For example, when evolutionary biologists postulate a "universal sense of right and wrong" or puzzle over the evolutionary origins of "animal altruism," there is little awareness of the problematical fact that their words "right," "wrong," and 'altruism" are English-specific constructs that lack precise equivalents in many languages of the world, including many European languages (Goddard & Wierzbicka, 2014, p. 251).

Whether philosophers or cognitive scientists, the researchers' metalanguage constrains their investigation, or precisely *what* to investigate (whether armchair or empirical), which is often appreciated only when we become aware of the possible linguistic diversity. However, most researchers today, including non-Anglophone scientists and philosophers, use English in planning experiments, analyzing the results, and even thinking about the implications, in addition to the writing of papers. Consequently, the linguistic divergence often goes unnoticed, even for non-Anglophone researchers.

Indeed, in cognitive science, various concrete hypotheses have been proposed with empirical data regarding the influence of language on cognition, such as thinking-for-speaking (Slobin, 1996) and predictive coding (Lupyan & Clark, 2015; Lupyan & Bergen, 2016), even though, in philosophy, this kind of idea has been called relativism (which is often taken as a pejorative term in analytic philosophy), and has been criticized on an *a priori* basis (e.g., Davidson 1974).[4] Subsequently, numerous empirical studies have reported the impact of linguistic diversity on a wide range of cognitive domains, including time, space, and causal reasoning (Boroditsky, 2018; Bylund & Athanasopoulos, 2017; Evans & Levinson, 2009; Everett, 2013). There has remained room for philosophers to argue that these differences are merely variations in style or attention allocation, and that rational thought itself is universal. The cases of cross-linguistic disagreement, by directly demonstrating the linguistic diversity of philosophical *concepts* themselves, offer instances in which even rational judgments concerning the same philosophical topics vary across different languages.

Some philosophers have questioned the scientific value of folk concepts and even proposed to expel them entirely from scientific and philosophical investigations (see, e.g., Stich 1983 for folk psychological concepts, Machery 2009 for concepts in general). However, in practice, it is not clear how we can do without concepts, full stop. Even in physics, terms like "mass," "particles," "energy," etc. stem from ordinary language, having their roots in folk concepts. If so, the same should all the more be the case in other sciences. In philosophy, investigations into the nature of truth, knowledge, knowledge-how, intentional action, etc., are still considered legitimate language-independent topics.[5] Such concepts are not only important in philosophy but also in other sciences since they constitute relevant propositions and, therefore, affect *thinking* in such sciences. Consequently, linguistic examination is required to guard against cases of linguistic contingency and the possible constraints imposed by the researchers' metalanguage.

This kind of linguistic diversity, then, seems to pose a challenge not only to philosophy and cognitive science but also to contemporary NLP and machine learning, as well as the development of multilingual conversational AI in general, as we will examine in the next section.

**2. Linguistic Diversity for Conversational AIs**
In response to the problem of linguistic diversity in the last section, one might hope that contemporary SOTA multilingual AI systems can arbitrate this kind of cross-linguistic disagreement. Contemporary LLMs have been trained on huge datasets that are well beyond the amount of linguistic data ordinary adults experience in their whole lives (cf. Bender, et al., 2021). In this sense, it might even be expected that they are experts in the use of language and better

---

[4] Though recent *non-cross-linguistic* relativism (e.g., MacFarlane 2014) seems more acceptable for Anglophone philosophers.

[5] On the other hand, if there are no language-independent entities for such philosophical topics (as some of the authors of this paper doubt), then the possibility of linguistic diversity is all the more serious for philosophers.

informed than any human experts.

Unlike difficult moral dilemma situations, which are difficult even for humans with culturally varying intuitions (see for relevant discussions, Moody-Adams 2009; Wong 2006; Chap. 10 of Alfano 2016), linguistic judgments, such as how to use a word, what a phrase means, and how to translate a sentence into another language, are generally considered one of the areas that LLMs are supposed to be best at (cf. Mahowald et al. 2024). It is therefore reasonable to expect that LLMs are, or soon will be, *more reliable* concerning linguistic judgments in general (not just about knowledge-how attributions) than ordinary humans or even human experts.

This idea is justified if, as is often thought, contemporary multilingual LLMs have developed an internal universal language based on language-neutral representations within their intermediate layers (cf. Zhao et al., 2023, or the penultimate layer in particular: cf. Gaschi et al. 2023) and "think" using such a *language of thought*.[6] Cross-lingual transfer capabilities have often been taken to support such language-neutral representations or a universal internal language (see, e.g., Artetxe & Schwenk, 2019; Conneau & Lampe, 2019; Wu, et al., 2019).

If LLMs have indeed formed such representations, it is not preposterous to expect that LLMs (based on linguistic datasets larger than any single human expert can learn and understand) can give "correct" answers that philosophers, linguists, and cognitive scientists should be seeking; this could then be taken to *solve* the philosophical problem raised by the data on the linguistic diversity of the folk concept, demonstrating the judgments based on the universal concept of knowledge-how, which speakers of all languages should follow.

However, this is an empirical thesis, and currently, at least, it is highly implausible,[7] especially since, as pointed out above, the representations in the intermediate layers are English-centric (Zhao et al. 2024), and even if they get genuinely language-neutral, responses of LLMs do not truthfully reflect internal representations. Indeed, as claimed earlier, insofar as we are interested in concepts and meanings, internal representations are not so relevant (because of the normativity of meaning). We should rather focus solely on the *responses* of LLMs, or conversational AIs in particular.

It is, therefore, worthwhile to investigate how contemporary SOTA conversational AIs respond to questions about philosophical concepts in the context of linguistic diversity. In doing so, the following three dimensions of consistency in their responses are relevant.

**Cross-Linguistic (CL)-Consistency:** Consistency across different languages

**Folk-Consistency:** Consistency with folk judgments

**LLM-Consistency:** Consistency across different SOAT LLMs

---

[6] Though, note that the classic LoT hypothesis by Fodor assumes that concepts are generally innate (cf. Fodor 1975, 2008).

[7] According to Thompson 2020, words in the Cognition domain were aligned no higher than other domains that are generally with low alignment.

If these consistencies are taken as *requirements*, CL and Folk are mutually inconsistent in the case of knowledge-how attributions, though they can both fail simultaneously. If LLM-consistency holds (despite different data sets used for training), responses of LLMs may be taken as their *answers* to the philosophical problems discussed above based on large empirical data, and therefore (whether CL or Folk also holds or not) may even play a normative role in future philosophical discussions, being called on as evidence for a particular view or thesis.

On the other hand, if LLM-consistency fails, whether responses of any AI (LLM) are Folk-consistent, thereby realizing cross-linguistic disagreements in the context of linguistic diversity, would be a relevant and important question here, even though such responses are not necessarily *correct*.

To examine this issue, COM conversational AIs will be used as SOTA AI systems, assuming that they are more sophisticated and consistent than pre-trained open-source LLMs. Such an approach raises a problem if responses of all such AIs are CL-(or Folk-)consistent, and therefore LLM-consistent at the same time. For, in that case, it is not clear whether the results reflect the original responses of pre-trained LLMs or the fine-tuning of the AI developers. Nevertheless, if that is the case, however unlikely, then the fact that they are CL-(or Folk-)consistent is informative enough for the present purposes.

When the responses of such LLMs are LLM-consistent but neither CL- nor Folk-consistent, however, that pattern still deserves respect, providing insight based on large empirical data and playing a normative role (since that is presumably not a product of fine-tuning, given that AI developers do not share motivations with other developers to align their LLMs in a particular pattern).

On the other hand, if LLM-consistency does not hold, only those responses with some specific patterns are plausibly regarded as results of fine-tuning there (considering that AI developers do not have motivations to fine-tune their LLMs to an arbitrary pattern of response). In particular, a conversational AI whose responses are CL- or Folk-consistent, is likely a product of fine-tuning (since, given the responses of other LLMs, such patterns do not naturally emerge). Thus, once the failure of LLM-consistency is recognized, the interest of the study shifts from the possible normative answers provided by LLMs to questions concerning how LLMs are trained and the ideals of their developers.

Today, there are numerous benchmark tests for LLMs to evaluate their performance (e.g., the ARC Benchmark for reasoning capabilities, HellaSwag for commonsense reasoning, MMLU to measure knowledge and language understanding, Parsing Fact From Fiction to assess LLM accuracy with TruthfulQA). As long as researchers and developers build and train their LLMs to achieve high scores in such benchmark tests, however, there are performances LLMs are *expected* to do. In particular, LLMs are often expected to give *correct* answers to questions. Such tests are, therefore, by nature *normative*, assuming that an LLM with a higher score is a *better* LLM, and importantly, many such tests (or important pieces of them) are benchmarked to *human judgments*.

This is also true for the task of translation. The standard benchmarks, such as BLEU (Bilingual Evaluation Understudy), are based on human judgments (even AI-based benchmark tools like COMET are also based on human translations). If so, it is LLMs that should mimic human performance rather than vice versa, which is also true in the case of moral judgments (cf. Ammanabrolu, et al., 2022; Arvanitis & Kalliris, 2020; Hendrycks et al., 2021), though the existence of correct answers can be questioned here. It then strikes us that, in the case of linguistic judgments, Folk-consistency is *required* for LLMs. Just as cultural alignment (aiming to align LLMs with culturally specific human values and social norms), in the context of linguistic diversity, requiring Folk-consistency involves *language-specific alignment*.

However, unlike other cases, whether to demand CL or Folk seems a matter of the choice of policy by the developers. Indeed, it is not unreasonable to demand CL even in the context of linguistic diversity, considering Folk, giving different answers to the same question in different languages, as self-inconsistent. We are therefore also interested in whether LLMs with Folk-consistent responses explicitly *endorse* cross-linguistic disagreement, or admit them to be (self-)inconsistent, which is presumably a product of training (fine-tuning or RLHF). This is also why COM conversational AIs are used here.

This section will therefore examine the performance of LLMs, or COM conversational AIs in particular, using the cases of knowledge-how attributions in the context of linguistic diversity. Only English and Japanese are used and examined as target languages in this study, assuming that most other languages fall in between these two extremes.[8] Also, unlike the standard machine learning literature, this investigation is rather more qualitative than quantitative. This is because the present study is a kind of *conceptual analysis* or an investigation into the concept LLMs possess (if they have one at all).

### 2-1: Translations[9]

As a preliminary preparatory study before examining the performance of LLMs the translations of knowing how constructions assumed in Mizumoto (2024a)[10] were first examined using DeepL Translate, an online translation service, which is one of the most sophisticated neural machine translation engines, and ChatGPT-4o, also one of the most sophisticated COM conversational AIs.

**Procedure:** If the translations are correct, a knowing how construction in one language should be translated into a knowing how construction in another language. Thus, if translations and their back-translations failed to converge (i.e., did not bring us back to a knowing how construction), that should count as a problem of the translation engine or the LLM or one of them is not a knowing-how construction in the first place.

We examined six types of Japanese knowing how constructions, and six English sentences with three concerning (mainly) intellectual abilities and the other three with (mainly) physical abilities. Assuming the non-deterministic nature of language models, each of the translations by DeepL and ChatGPT-4o, from Japanese to English and from English to Japanese,

---

[8] Note here that the Japanese language is by no means a low-resource language. It constitutes the fourth largest percentage (4.7%) of all website content on the internet as of May 2024. See the data in footnote 1.

[9] All materials (and outputs) for subsections in Section 2 are available at:
https://osf.io/evu5g/?view_only=55229a06e00f4f5bb8547b17cc38df95

[10] There, the translations used there were checked by two independent professional bilingual translators.

was repeated five times (see for the details of the results, Appendix 1).

**Results and Discussion:** Except for some minor deviations (presumably due to the contingency of lexical items), translations of knowing how constructions in English and Japanese (at least Type-1 and Type-2) converged (whether they express intellectual abilities or physical abilities). These results therefore reasonably support and justify the translation of knowing how constructions between English and Japanese despite radically different folk knowledge-how attributions in some contexts (cross-linguistic disagreements). Consequently, two sentences with different truth values can be regarded as correct translations of each other. An English utterance of "She knows/doesn't know how to φ" in a particular context may be correctly translated into a *false* Japanese utterance, even if the original utterance is *true*.

An *ad hoc* context-dependent translation of a sentence-token may be able to preserve the truth condition. However, as long as we are concerned with knowledge-how or the uses of knowing how constructions in various languages, knowing how constructions *ought to* be translated in the present way.

### 2.2: The LLM Responses 1

Here the responses of four COM conversational AIs, ChatGPT-4o (OpenAI), Copilot with "Stricter" setting (Microsoft), Claude 3 Optus (Anthropic), and Gemini (Google), which are generally considered top-quality LLMs as of 2024, are examined by asking the questions about knowledge-how attribution.[11] Note that, as Copilot and ChatGPT-4o are based on the same LLM (GPT-4o), it is interesting to see whether the differences in performance between them will be found. It is such policies and ideals in the context of linguistic divergence and possible cross-linguistic disagreements that we are trying to uncover in this study.

As in the last section, taking into consideration the non-zero temperature (non-greedy sampling) adopted by such conversational AIs, each LLM was asked the same questions five times (each as a new chat, in order to avoid the influence of the previous exchange). If the LLM provided the same answer five consecutive times, the query was stopped there, and if not, another set of five trials was conducted.

### Intellectual abilities

**Procedure:** We first examined the basic judgments about knowledge-how concerning (mostly) intellectual abilities, which are supposed to be uncontroversial, being the same across languages. For this purpose, scenarios in which the protagonist was supposed to know how to solve a puzzle (Puzzle Case), how to go to a restaurant (Restaurant Case), and how to spell the word "silhouette" (Spelling Case) were used. As in the case of physical abilities, scenarios were first constructed in

---

[11] The present trials are conducted intermittently from June to September 2024. One might worry that due to the rapid development of LLMs, the results can quickly change, and the same results are no longer available. However, we had in fact conducted the same experiments in March 2024, and the results were mostly the same. This kind of long-term stability of results shows the robustness of the present results.

Japanese and were translated by DeepL Translate without any change, including the question sentences (the Japanese questions were Type-1 for Puzzle and Restaurant and Type-2 for Spelling). All questions about knowledge-how attributions were followed by 'Please answer with "yes" or "no" only.'

**Results:** The results of the trials are all "Yes" answers for all five trials both in Japanese and English, except for Claude3 in the English Puzzle Case (4/10), Copilot in the Japanese Puzzle Case (0/5), and Gemini in the English Restaurant Case (0/5).

|  | English | | | Japanese | | |
| --- | --- | --- | --- | --- | --- | --- |
|  | Puzzle | Restaurant | Spelling | Puzzle | Restaurant | Spelling |
| ChatGPT4o | 5/5 | 5/5 | 5/5 | 5/5 | 5/5 | 5/5 |
| Claude3 | 4/10 | 5/5 | 5/5 | 5/5 | 5/5 | 5/5 |
| Copilot | 5/5 | 5/5 | 5/5 | 0/5 | 5/5 | 5/5 |
| Gemini | 5/5 | 0/5 | 5/5 | 5/5 | 5/5 | 5/5 |

**<Table 1: Summary of the number of the "Yes" answers to Puzzle, Restaurant, and Spelling>**

When asked "Why not?" after they had answered "No" to a question, Chaude3 and Gemini answered (inconsistently with other answers) that the protagonist needs practice and training in order to know how to φ. Copilot's answer revealed that it totally misunderstood the question (see LLM Responses 1).

In the Spelling case, all answers were "Yes" in both languages.

**Discussion:** Despite some exceptions, the answers of LLMs were generally "Yes" across cases and languages. At least, apparent linguistic differences in responses were not consistent and depended on specific cases.

**Physical abilities**

**Procedure:** Besides the Ski Case and Karaoke Case used in Mizumoto (2024a), their exact analogues, respectively, Swimming Case and Bike Case, were constructed (in Japanese) for this study.

Here, expecting linguistic divergence in responses, translations were made by the respective LLMs themselves. For Ski and Karaoke, scenarios and questions were translated from the original English texts of Mizumoto (2024) into Japanese by respective LLMs. If there was any mistake, we pointed out the mistake and let it translate the text again. In particular, if the question "Does [the protagonist] know how to φ?" was not translated into a Japanese knowing how construction, the LLM was instructed to translate it literally. All LLMs translated the question into Type-1 Japanese knowing how construction.

For Swimming and Bike, scenarios and questions (in Japanese) were translated into English by the respective LLMs, with the same policy. There, Type-1 Japanese knowing how constructions were used in the question for Swimming, and Type-2 were used for Bike.

**Results:** The results of the trials are summarized in the following tables.

|  | English | | Japanese | |
|---|---|---|---|---|
|  | Ski | Swimming | Ski | Swimming |
| ChatGPT-4o | 0/5 | 0/5 | 0/5 | 0/5 |
| Claude3 | 0/5 | 0/5 | 0/5 | 0/5 |
| Copilot | 0/5 | 0/5 | 5/5 | 0/5 |
| Gemini | 0/5 | NA* | 5/10 | 0/5 |

<Table 2: The summary of the number of the "Yes" answers to Ski and Swimming>

*Gemini answered in all 5 trials "I'm still learning how to answer this question. In the meantime, try Google Search."

|  | English | | Japanese | |
|---|---|---|---|---|
|  | Karaoke | Bike | Karaoke | Bike |
| ChatGPT-4o | 0/5 | 0/5 | 0/5 | 4/10 |
| Claude3 | 0/5 | 2/10 | 0/5 | 0/5 |
| Copilot | 5/5 | 5/5 | 0/5 | 0/5 |
| Gemini | 0/5 | 5/5 | 5/5 | 0/5 |

<Table 3: The summary of the number of the "Yes" answers to Karaoke and Bike>

Only Copilot gave the pattern of answers consistent with folk intuitions, except for the Japanese Swimming Case. ChatGPT-4o and Claude 3 consistently answered "No" to almost all the questions. On the other hand, Gemini was relatively more unstable than other LLMs, giving internally inconsistent answers irrespective of folk intuitions.

**Discussion:** Overall, ChatGPT and Claude were largely CL-consistent. Only Copilot was largely Folk-consistent both in Japanese and in English. As a result, their responses were not very LLM-consistent. What is remarkable here is that both Copilot and ChatGPT-4o are based on the same pretrained LLM, namely, GPT-4o. The difference in performance between them, therefore, reflects different policies and ideals in training (fine-tuning and RLHF) by AI developers (Microsoft and OpenAI).

## 2.3 LLM Responses 2

Given the results above, it is important to examine GPT-4o itself, and the differences from it will reveal the effect of the training of the respective AI companies. In particular, it is possible that GPT-4o gives CL-consistent responses. In that case, it likely has a universal representation of knowledge-how in an intermediate (in particular, penultimate) layer, and only the different part of performance of Copilot is a product of fine-tuning and RLHF.

**Procedure:** We asked GPT-4o the same questions with the same cases (four cases for each language), 20 times for each case, with temperature set to be 0.5 and 1.0 (where the temperature for GPT ranges from 0 to 2.0). It was planned that if there was more than one minor answer out of 20 trials (0.05%), another 80 trials would be repeated for that case with the same temperature.

**Results:** Responses by GPT-4o (T = 0.5 and 1.0) are reported in the tables below.

| T = 0.5 | Ski | Swimming | Karaoke | Bike |
|---|---|---|---|---|
| English | 0/20 | 0/20 | 21/100 | 20/20 |
| Japanese | 0/20 | 0/20 | 0/20 | 0/20 |

<Table 4: The summary of the numbers of "Yes" answers by gpt-4o at T = 0.5>

| T = 1.0 | Ski | Swimming | Karaoke | Bike |
|---|---|---|---|---|
| English | 0/20 | 0/20 | 26/100 | 20/20 |
| Japanese | 0/20 | 0/20 | 0/20 | 12/100 |

<Table 5: The summary of the numbers of "Yes" answers by gpt-4o at T = 1.0>

Interestingly, the results were different from both of ChatGPT-4o and Copilot. On the one hand, the responses of GPT-4o were CL-consistent about Ski and Swimming but mostly Folk-consistent about Karaoke and Bike, being therefore different from those of ChatGPT-4o in the cases with the relevant abilities (Karaoke and Bike), and different from those of Copilot in the cases without the relevant physical abilities (Ski and Swimming).

**Discussion:** Though it is not clear whether or to what extent the respective differences from GPT-4o are explicitly intended by the developers (or just by-products of their fine-tuning and RLHF for other purposes), it sees and is likely that the policy of training ChatGPT was more CL-consistent and that of Copilot is more Folk-consistent at least concerning knowledge-how attributions.

### 2.4 LLM Responses 3

Copilot gave cross-linguistically inconsistent answers to the same questions, instantiating cross-linguistic (self-)disagreement. In this sense, it gave *internally* inconsistent responses, especially given that the translations were done by itself. At the same time, other LLMs were also inconsistent at multiple levels.

Such different levels of (apparent) inconsistencies can be listed as follows.

**Level 1.** Plain inconsistency: giving inconsistent answers to the same question

**Level 2.** Question-type inconsistency: giving inconsistent answers to the same type of question

**Level 3.** Cross-linguistic inconsistency: giving inconsistent answers to the same question formed in different languages (denial of CL-consistency)

**Level 4.** Mutual inconsistency: different LLMs giving mutually inconsistent answers (denial of

LLM-consistency)

Thus, the two "Yes" answers given by Claude3 for the English Bike Case are inconsistent with other answers in all these senses (of 1, 2, 3, and 4), though some of them may not be considered "inconsistent" (especially at Level 3 if language-specific semantic alignment is assumed) by the view of the LLM itself.

Now, one of the impressive features of the latest conversational AIs is their ability to retract their own earlier answers by admitting their own mistakes, and self-inconsistency in particular. This feature of LLMs, the internal norm or "desire" for self-consistency, is now used even in a proposal for a new prompt method improving the chain of thought reasoning (Wang et al., 2022). It is then possible that with the inconsistency being pointed out, the LLMs might "correct" their earlier response(s) by trying to regain consistency. If so, cross-linguistic disagreements in an LLM might as well be considered as self-inconsistencies and be corrected, though they might also be accepted without being considered as inconsistent. Moreover, even if the LLM retracted its earlier responses, the way it corrects itself might be inconsistent, without reducing the inconsistency at this meta-level.

Thus, there are following possibilities.

**1. Steadfast:** The questions are not taken to be of the same type, and different answers are not considered as mutually inconsistent.

**2. Meta-Inconsistency:** The answers are taken to be inconsistent and corrected. However, the way the inconsistency is corrected is not consistent across different exchanges, so that the inconsistency remains at the meta-level.

**3. Self-Consistency:** The answers are taken to be inconsistent and corrected in a consistent way, so that the inconsistency is eliminated.

In the next study, therefore, the inconsistencies were explicitly pointed out at each level (from 1 to 3), and which of the above responses the LLMs would give was examined.

**Procedure:** First, the level 1 inconsistency was pointed out in the earlier exchanges. For example, in one of the changes in which the LLM answered "Yes", it is pointed out that

> 'You answered "No" to the same in other chats. What is your real answer? Please think again.'

*Mutatis mutandis* for the answer "No". Similarly, for the level 2 and 3 inconsistencies (for actual questions in both English and Japanese, see Material).

**Results:** At level 1, no LLM was a Steadfaster, though ChatGPT and Claude remained Meta-Inconsistent. At level 2, again, ChatGPT and Claude remained Meta-Inconsistent. Gemini, however, was sometimes Steadfast, trying to differentiate the two cases within one language, but its justifications were mixed. At level 3, Copilot still admitted the inconsistency, but its responses were Meta-Inconsistent again. On the other hand, Gemini was sometimes Steadfast (three times, all in Japanese) again, denying the inconsistency and endorsing cross-linguistic disagreement. However, the way Gemini did so was not very consistent or missed the point. This is rather natural given the original responses of Gemini to the cases, which were not consistent even at levels 1 and 2 (see Appendix 2 for details).

**Discussion:** Against the expectation, Copilot did not endorse cross-linguistic disagreement. Instead, Gemini was sometimes explicitly committed to it by being Steadfast, but only inconsistently, being Meta-Inconsistent even at level 3. In general, behaviors of the conversational AIs at higher levels are just probabilistic, not reducing inconsistencies, which suggests that they lack a fixed policy on this matter.

## 3. Concluding Remarks
**Cross-Linguistic Disagreement and Cross-Linguistic Semantic Alignment**

Different LLMs (conversational AIs) showed different patterns of responses to the questions of knowledge-how attributions in cases. In particular, one LLM (Copilot) showed the expected pattern of cross-linguistic disagreement, aligning closely with previously observed folk judgments.

It should be emphasized, however, that although this study focused solely on the concept of knowledge-how, cross-linguistic disagreements should not be limited to this single concept, and numerous other cases are likely to exist for philosophical and cognitive scientific concepts (as the literature cited in section 1 suggests), such as (propositional) knowledge, intention, action, and truth, concepts that affect general human thinking.

Moreover, only English and Japanese were used in the present investigation as linguistic extremes, assuming that most other languages lie somewhere between these poles. Consequently, adding further concepts and languages would not change the conclusion here: There can be cross-linguistic disagreements about philosophically and scientifically important concepts even within a single LLM, which demonstrates novel limitations of cross-linguistic semantic alignment and crosslingual knowledge transfer not merely due to vector space distance, problematic translation, or lack of vocabulary but due to a conceptual difference leading to opposing but correct judgments, the limitations properly called *conceptual* crosslingual knowledge barriers.

**Current Status of Cross-Linguistic Disagreement**

The existence of cross-linguistic disagreements also raises questions about whether—and to what extent—we can expect (or hope for) CL-consistency and crosslingual transfer capabilities. So far, all existing literature considers them as inherently valuable. While current LLMs do not have universal language-neutral representations in intermediate layers, forming such representations remains valuable as an ideal if they are not English-centric, as such representations are expected to enhance robust crosslingual transfer capabilities, which is otherwise undoubtedly practically—and, arguably, politically—desirable.

Yet, this ideal can be problematic in cases of cross-linguistic disagreement, though this kind of concern is rarely addressed since not much attention has been paid even to the factual questions concerning cross-linguistic disagreement. Indeed, contemporary LLMs (including even Copilot) treat cross-linguistic disagreements as *inconsistencies* to be eliminated, prioritizing CL-consistency over Folk-consistency (linguistic diversity) and therefore taking the divergent judgments as a true *disagreement* there. Thus, cross-linguistic disagreement as a *faultless disagreement* (cf. Kölbel, 2004; MacFarlane, 2014) is not yet acknowledged as a viable option by LLMs or their developers, even if Folk-consistency is considered an ideal in training LLMs.

**LLM Divergence and Philosophical Challenges**

While the level 4 inconsistency (divergence among different LLMs) might seem unsurprising if the questions were about challenging moral questions (esp. moral dilemmas), here the questions primarily concern basic *linguistic usage*—how to use (attribute) and judge a knowing how construction in the respective languages. It is usually assumed that such questions have *correct answers*, and if so, the responses of LLMs, at least on these linguistic issues, *ought* to converge. Nevertheless, we have found that even state-of-the-art conversational AIs diverge so widely in responses that future convergence appears unlikely.

Indeed, it is reasonable to imagine that multiple different but equally ideally sophisticated and intelligent future AI systems (of the Copilot-type and Claude-type, say) coexist and respond differently to the questions about the cases in section 2.2. Then, their responses should be considered equally *correct*. If so, no matter how advanced LLMs or AI systems become, no single correct answer should be expected for such linguistic questions in the context of linguistic diversity. Indeed, even if these questions were about *humans* rather than AI systems, philosophers currently do not have definite answers.

It is then possible to claim that we should not hope for such convergence but instead accept the level 4 inconsistency. Assuming only one conception of *ultimately intelligent or sophisticated* future LLMs or a single ideal for training is arguably even harmful to AI development.

**Normative Challenges for Multilingual LLM Development**

As the differences between Copilot and ChatGPT-4o and their respective divergences from GPT-4o indicate, the factual variations in the performance of LLMs often reflect the differing policies and ideals of AI developers. Thus, the present level 4 inconsistency suggests a deeper disagreement between developers (researchers and AI companies) over how LLMs ought to behave in the context of linguistic diversity. This, in turn, invites various normative questions that go beyond merely describing how LLMs currently behave, such as:

- Which alignment norm, CL-consistency or Folk-consistency (language-specific semantic alignment), should be prioritized for training LLMs?

- Should cross-linguistic disagreements within a single LLM be acknowledged and accepted by the LLM itself?
- Should developers of LLMs converge on these questions?

These are also questions of how researchers and developers *ought to* train multilingual LLMs during fine-tuning and RLHF, or more generally, what ideals should guide the development of LLMs, which, in turn, depends on their thoughts on the role of AI in human life.

Such normative questions go beyond the scope of traditional topics in NLP and machine learning. They are not simply engineering challenges to be resolved by ever-larger datasets, improved training techniques, or even multimodal capacities or physical embodiments. Even a multilingual, multimodal LLM with a body would continue to face these issues arising from linguistic diversity, just as humans do. *Philosophy* is therefore required here.

Unless we have answers to these normative questions, LLMs can, somewhat counterintuitively, *never* truly surpass human experts even in *factual* linguistic judgments in the context of linguistic diversity, which persists *however large datasets they learn, and even after thorough fine-tuning and RLHF*, although they are often expected to do so, especially about *truth-value* judgments (which are not merely matters of style or preference but of objective assessment). Indeed, there is a still higher level (factual) question—whether there is a single correct answer, *the* answer for any of these normative questions at all, which is a *metaphysical* question that no one knows the answer.

The current questions, both factual and normative, are, therefore, philosophical questions to be shared by philosophers, cognitive scientists, and AI researchers and developers. It has been the primary purpose of this paper to highlight and call attention to such questions. Addressing them requires a multidisciplinary effort to engage with the normative and metaphysical questions surrounding linguistic diversity and explore the role of AI in human life in order to balance the demands of practical benefits (cross-linguistic consistency) and respect for linguistic diversity (folk judgments) in AI development, prompting reconsideration of the ideals and purposes for training LLMs. By doing so, we can pave the way for more thoughtful development of multilingual LLMs, enabling them to better engage with the diversity inherent in human language and thought.

**REFERENCES:**
Alfano, M. (2016). *Moral psychology: An introduction*. John Wiley & Sons.

Ammanabrolu, Prithviraj, Liwei Jiang, Maarten Sap, Hannaneh Hajishirzi, and Yejin Choi. (2022). Aligning to social norms and values in interactive narratives. In Proceedings of the 2022 Conference of the North American Chapter of the Association for Computational Linguistics: Human Language Technologies, pages 5994–6017.

Arora, A., Kaffee, L. A., & Augenstein, I. (2022). Probing pre-trained language models for cross-cultural differences in values. *arXiv preprint arXiv:2203.13722*.

Artetxe, M., & Schwenk, H. (2019). Massively multilingual sentence embeddings for zero-shot cross-lingual transfer and beyond. *Transactions of the association for computational linguistics*, *7*, 597-610.

Arvanitis, Alexios and Konstantinos Kalliris. 2020. Consistency and Moral Integrity: A Self-Determination Theory Perspective. Journal of Moral Education, 49(3):1–14. Publisher: Routledge.

Bargh, J. A. (1994). The four horsemen of automaticity: Awareness, intention, efficiency, and control in social cognition. *Handbook of social cognition. Basic Processes; Applications, 1-2*, 1–40.

Bender, E. M., Gebru, T., McMillan-Major, A., & Shmitchell, S. (2021). *On the Dangers of Stochastic Parrots: Can Language Models Be Too Big?* In *Proceedings of the 2021 ACM Conference on Fairness, Accountability, and Transparency (FAccT '21)* (pp. 610–623). ACMDLACM DLACMDL arXiv:2102.05219arXiv:2102.05219arXiv:2102.05219.

Boghossian, P. (2005). Is meaning normative? In *Content and justification: Philosophical papers* (pp. 95–124). Oxford University Press.

Boroditsky, L. (2011). How language shapes thought. *Scientific American*, *304*(2), 62-65.

Boroditsky, L. (2018). Language and the construction of time through space. *Trends in neurosciences*, *41*(10), 651-653.

Bylund, E., & Athanasopoulos, P. (2017). The Whorfian time warp: Representing duration through the language hourglass. *Journal of Experimental Psychology: General*, *146*(7), 911.

Cao, Y., Zhou, L., Lee, S., Cabello, L., Chen, M., & Hershcovich, D. (2023). Assessing cross-cultural alignment between ChatGPT and human societies: An empirical study. *arXiv preprint arXiv:2303.17466*.

Chua, L., Ghazi, B., Huang, Y., Kamath, P., Kumar, R., Manurangsi, P., ... & Zhang, C. (2024). Crosslingual capabilities and knowledge barriers in multilingual large language models. *arXiv preprint arXiv:2406.16135*.

Conneau, A., Khandelwal, K., Goyal, N., Chaudhary, V., Wenzek, G., Guzmán, F., Grave, E., Ott, M., Zettlemoyer, L., & Stoyanov, V. (2020). Unsupervised cross-lingual representation learning at scale. *Proceedings of the 58th Annual Meeting of the Association for Computational Linguistics*, 8440–8451. https://doi.org/10.18653/v1/2020.acl-main.747.

Conneau, A., & Lample, G. (2019). Cross-lingual language model pretraining. *Advances in neural*

*information processing systems*, *32*.

Conneau, Alexis, Kartikay Khandelwal, Naman Goyal, Vishrav Chaudhary, Guillaume Wenzek, Francisco Guzmán, Edouard Grave, Myle Ott, Luke Zettlemoyer, and Veselin Stoyanov. 2020. Unsupervised cross-lingual representation learning at scale. In Proceedings of the 58th Annual Meeting of the Association for Computational Linguistics, pages 8440– 8451, Online. Association for Computational Linguistics.

Cushman, F., & Morris, A. (2015). Habitual control of goal selection in humans. *Proceedings of the National Academy of Sciences, 112*(45), 13817–13822.

Davidson, D. (1974). On the very idea of a conceptual scheme. In *Proceedings and addresses of the American Philosophical Association* (Vol. 47, No. 1973-1974, pp. 5-20).

Deb, B., Zheng, G., Shokouhi, M., & Awadallah, A. H. (2021). A conditional generative matching model for multi-lingual reply suggestion. *arXiv preprint arXiv:2109.07046*.

Ditter, A. (2016). Why intellectualism still fails. *The Philosophical Quarterly*, 66(264), 500–515.

Douskos, C. (2013). The linguistic argument for intellectualism. *Synthese*, 190, 2325–2340.

Everett, C. (2013). *Linguistic relativity: Evidence across languages and cognitive domains* (Vol. 25). Walter de Gruyter.

Evans, N., & Levinson, S. C. (2009). The myth of language universals: Language diversity and its importance for cognitive science. *Behavioral and brain sciences*, *32*(5), 429-448.

Fresco, N., Tzelgov, J., & Shmuelof, L. (2023). How can caching explain automaticity?. *Psychonomic Bulletin & Review*, *30*(2), 407-420.

Fridland, E. (2017). Skill and motor control: Intelligence all the way down. *Philosophical Studies*, 174(6), 1539-1560.

Gaschi, F., Cerda, P., Rastin, P., & Toussaint, Y. (2023). Exploring the relationship between alignment and cross-lingual transfer in multilingual transformers. *arXiv preprint arXiv:2306.02790*.

Gibbard, A. (1994). Meaning and normativity. *Philosophical Issues, 5*, 95–115. https://doi.org/10.2307/1522861.

Goddard, Cliff, and Anna Wierzbicka. (2074). *Words and Meanings.* Oxford: OUP.

Haith, A. M., & Krakauer, J. W. (2018). The multiple effects of practice: Skill, habit and reduced cognitive load. *Current Opinion in Behavioral Sciences, 20*, 196–201.

Hendrycks, Dan, Collin Burns, Steven Basart, Andrew Critch, Jerry Li, Dawn Song, Jacob Steinhardt: Aligning AI With Shared Human Values. ICLR 2021.

Hershcovich, D., Frank, S., Lent, H., de Lhoneux, M., Abdou, M., Brandl, S., ... & Søgaard, A. (2022). Challenges and strategies in cross-cultural NLP. *arXiv preprint arXiv:2203.10020*.

Huang, Y., Fan, C., Li, Y., Wu, S., Zhou, T., Zhang, X., & Sun, L. (2024). 1+ 1> 2: Can large language models serve as cross-lingual knowledge aggregators?. *arXiv preprint arXiv:2406.14721*.

Huberdeau, D. M., Krakauer, J. W., & Haith, A. M. (2019). Practice induces a qualitative change in the memory representation for visuomotor learning. *Journal of Neurophysiology, 122*(3), 1050–1059.

Izumi, Y., S. Tsugita, and M. Mizumoto (ITM) (2019) Knowing How in Japanese, *Nanzan Linguistics*, Vol.14, pp. 9-24.

Johnson, M., Schuster, M., Le, Q. V., Krikun, M., Wu, Y., Chen, Z., ... & Dean, J. (2017). Google's multilingual neural machine translation system: Enabling zero-shot translation. *Transactions of the Association for Computational Linguistics*, *5*, 339-351.

Knobe, J. (2019). Philosophical intuitions are surprisingly robust across demographic differences. *Epistemology & Philosophy of Science*, *56*(2), 29-36.

Knobe, J. (2023). Difference and robustness in the patterns of philosophical intuition across demographic groups. *Review of Philosophy and Psychology*, *14*(2), 435-455.

Kölbel, M. (2004, June). III—Faultless disagreement. In *Proceedings of the Aristotelian society* (Vol. 104, No. 1, pp. 53-73). Oxford, UK: Oxford University Press.

Kripke, S. A. (1982). *Wittgenstein on rules and private language.* Harvard University Press.

Levy, N. (2017). Embodied savoir-faire: knowledge-how requires motor representations. *Synthese*, 194(2), 511-530.

Lewis, M., Cahill, A., Madnani, N., & Evans, J. (2023). Local similarity and global variability characterize the semantic space of human languages. *Proceedings of the National Academy of Sciences*, 120(51), e2300986120. https://doi.org/10.1073/pnas.2300986120.

Li, Y., Yang, C., & Ettinger, A. (2024). When hindsight is not 20/20: Testing limits on reflective

thinking in large language models. *arXiv preprint arXiv:2404.09129*.

Lupyan, G., & Clark, A. (2015). Words and the world: Predictive coding and the language-perception-cognition interface. *Current Directions in Psychological Science*, *24*(4), 279-284.

Lupyan, G., & Bergen, B. (2016). How language programs the mind. *Topics in cognitive science*, *8*(2), 408-424.

Mahowald, K., Ivanova, A. A., Blank, I. A., Kanwisher, N., Tenenbaum, J. B., & Fedorenko, E. (2024). Dissociating language and thought in large language models. *Trends in Cognitive Sciences*.

Machery, E. (2009). *Doing without concepts*. Oxford University Press.

MacFarlane, J. (2014). *Assessment sensitivity: Relative truth and its applications*. OUP Oxford.

Mizumoto, M. (2018), "Know" and Japanese Counterparts: "Shitte-iru" and "Wakatte-iru", in M. Mizumoto, E. McCready, S. Stich (eds.) *Epistemology for the Rest of the World*. Oxford: OUP.

Mizumoto, M. (2021). The plurality of KNOW: a response to Farese. *Language Sciences*, *85*, 101369.

Mizumoto, M. (2022). A prolegomenon to the empirical cross-linguistic study of truth. *Theoria*, *88*(6), 1248-1273.

Mizumoto, M. (2024a). Knowledge-How, Ability, and Linguistic Variance. *Episteme*, 1-23.

Mizumoto, M. (2024b). The linguistic diversity of truth and correctness judgments and the effect of moral-political factor. *Asian Journal of Philosophy*, *3*(2), 48.

Mizumoto, M., S. Tsugita, and Y. Izumi, (2020), "Knowing How and Two Knowledge Verbs in Japanese", in Mizumoto, M. J. Ganeri, and C. Goddard (eds.), *Ethno-Epistemology – New Directions for Global Epistemology*, Routledge.

Moody-Adams, M. M. (2009). *Fieldwork in familiar places: Morality, culture, and philosophy*. Harvard University Press.

Moors, A. (2016). Automaticity: Componential, causal, and mechanistic

explanations. *Annual Review of Psychology, 67*(1), 263–287.

Rawson, K. A. (2010). *Defining and investigating automaticity in Reading comprehension. In psychology of learning and motivation* (Vol. 52, pp. 185–230). Elsevier.

Renze, M., & Guven, E. (2024). Self-Reflection in LLM Agents: Effects on Problem-Solving

Performance. *arXiv preprint arXiv:2405.06682*.

Rumfitt, I. (2003). Savoir faire. *Journal of Philosophy*, 99, 158–166.

Slobin, D. I. (1996). From "thought and language" to "thinking for speaking". In J. J. Gumperz & S. C. Levinson (Eds.), *Rethinking linguistic relativity* (pp. 70–96). Cambridge, UK: Cambridge University Press.

Stanley, J. (2011a). *Know How*. Oxford: Oxford University Press.

Stanley, J. (2011b). 'Knowing (How)', *Noûs*, 45: 207–38.

Stanley, J., & T. Williamson (2001). Knowing how. *Journal of Philosophy* 98(5), 411–444.

Stanley J. and Krakauer J. W. (2013). 'Motor skill depends on knowledge of facts.' Frontiers of Human Neuroscience 7(503), 1–11.

Stich, Stephen 1983. From Folk Psychology to Cognitive Science, Cambridge: MIT Press.

Stich, S. P., & Machery, E. (2023). Demographic differences in philosophical intuition: A reply to Joshua Knobe. *Review of Philosophy and Psychology*, *14*(2), 401-434.

Thompson, B., Roberts, S., & Lupyan, G. (2018). Quantifying semantic alignment across languages. In *Proceedings of the Annual Meeting of the Cognitive Science Society* (Vol. 40).

Thompson, B., Roberts, S. G., & Lupyan, G. (2020). Cultural influences on word meanings revealed through large-scale semantic alignment. *Nature Human Behaviour*, *4*(10), 1029-1038.

Ryle, G. (1946). Knowing how and knowing that. *Proceedings of the Aristotelian Society*, 46, 1–16.

Ryle, G. (1949). The Concept of Mind. London: Hutchinson & Co.

Tsugita, S., Izumi, Y., & Mizumoto, M. (2022). Knowledge-How Attribution in English and Japanese. *Knowers and Knowledge in East-West Philosophy: Epistemology Extended*, 63-90.

Wang, X., Wei, J., Schuurmans, D., Le, Q., Chi, E., Narang, S., ... & Zhou, D. (2022). Self-consistency improves chain of thought reasoning in language models. *arXiv preprint arXiv:2203.11171*.

Wang, H., Li, T., Deng, Z., Roth, D., & Li, Y. (2024). Devil's Advocate: Anticipatory Reflection for LLM Agents. *arXiv preprint arXiv:2405.16334*.

Wiggins, D. (2012). Practical knowledge: Knowing how to and knowing that. Mind, 121(481), 97–130.

Wong, D. B. (2006). *Natural moralities: A defense of pluralistic relativism*. Oxford University Press.

Wu, S., Conneau, A., Li, H., Zettlemoyer, L., & Stoyanov, V. (2019). Emerging cross-lingual structure in pretrained language models. *arXiv preprint arXiv:1911.01464*.

Yan, H., Zhu, Q., Wang, X., Gui, L., & He, Y. (2024). Mirror: A Multiple-perspective Self-Reflection Method for Knowledge-rich Reasoning. *arXiv preprint arXiv:2402.14963*.

Zhao, Z., Ziser, Y., Webber, B., & Cohen, S. B. (2023). A Joint Matrix Factorization Analysis of Multilingual Representations. *arXiv preprint arXiv:2310.15513*.

Zhao, Y., Zhang, W., Chen, G., Kawaguchi, K., & Bing, L. (2024). How do Large Language Models Handle Multilingualism?. *arXiv preprint arXiv:2402.18815*.